\documentclass[13pt]{article}
\usepackage{graphicx}
\usepackage{amsmath}
\usepackage[hidelinks]{hyperref}
\usepackage{geometry}
\usepackage{float}  

\title{Simulation-Based Control Architecture Using Webots and Simulink}
\author{
Harun KURT\textsuperscript{1} Ahmet CAYIR\textsuperscript{1} \\
$^{1}$Yıldız Technical University, Department of Mechatronics Engineering \\
Istanbul, Turkey \\
\texttt{harun.kurt@std.yildiz.edu.tr} \texttt{ahmet.cayir@std.yildiz.edu.tr}
}

\date{Supervisor: Prof. Kadir ERKAN\textsuperscript{1} \\ \texttt{kerkan@yildiz.edu.tr} \\ April 16, 2025}

\begin{document}

\maketitle
\begin{abstract}
This paper presents a simulation based control architecture that integrates Webots and Simulink for the development and testing of robotic systems. Using Webots for 3D physics based simulation and Simulink for control system design, real time testing and controller validation are achieved efficiently. The proposed approach aims to reduce hardware in the loop dependency in early development stages, offering a cost effective and modular control framework for academic, industrial, and robotics applications.
\end{abstract}

\section{Introduction}
Simulation environments play a crucial role in the design, testing, and validation of control architectures for robotic systems. Webots, a powerful open-source robot simulator, provides a realistic physics engine and graphical interface for mobile and industrial robots. On the other hand, MATLAB/Simulink is widely used for model-based design and real-time control system implementation.\\

Integrating these two platforms enables engineers to rapidly prototype and test control algorithms before deploying them to physical hardware. This integration is particularly useful in academic research, where iterative testing is frequent and access to physical systems may be limited.

\section{System Architecture}
\subsection{Webots Environment}
Webots is used to model the robot, including its kinematics, dynamics, and sensors (e.g., IMU, LIDAR, encoders). The environment also simulates physical interactions such as collisions and friction. A robot controller, written in C, Python, or MATLAB, handles data transmission between Webots and external interfaces.

\subsection{Simulink Controller}
The control logic is implemented in Simulink, enabling rapid development of feedback controllers (PID, LQR, MPC, etc.). Sensor data from Webots is transmitted to Simulink via UDP or shared memory blocks, where it is processed and used to compute actuator commands. These commands are then sent back to Webots in real time.

\subsection{Simulink-Webots Binding}

The architectural binding between MATLAB/Simulink and Webots is established through custom interfaces that allow seamless data exchange in real time. In this setup, Webots simulates the physical robot and environment, while Simulink operates as the controller design and execution platform.\\

There are two primary ways of binding Webots and Simulink:

\begin{itemize}
    \item \textbf{1 External Communication via UDP or TCP:} This method involves writing custom Webots robot controller code (in Python, C, or MATLAB) to send sensor data to Simulink via UDP sockets. Simulink receives the data using the UDP Receive block and processes it to generate actuator commands, which are then sent back using the UDP Send block. This method offers flexibility but introduces latency that can affect high-frequency control.
    
    \item \textbf{2 Simulink C MEX S-Functions or Webots-MATLAB API:} A more integrated and efficient approach is to develop S-Function blocks or to use the Webots-MATLAB interface provided via Matlab [3]. These methods allow synchronous communication and real-time simulation without relying on network protocols.
\end{itemize}

In this study, the second method (Simulink-Webots direct integration using custom S-Functions or Webots-MATLAB API) is adopted. Sensor readings from the Webots simulation are transferred directly to Simulink through custom-designed interface blocks, where control logic is executed. The generated control outputs are then returned to Webots in real time.\\

This direct architectural binding eliminates the overhead of external communication protocols and significantly enhances simulation performance, especially for high-speed feedback loops. Moreover, it allows for rapid development, debugging, and real-time testing of control algorithms in a modular and scalable manner, making it highly suitable for academic and research applications.

\subsection{Communication Interface}
The communication between Webots and Simulink is typically established using:
\begin{itemize}
  \item \textbf{Shared memory:} For higher-speed interaction in local setups.
  \item \textbf{MATLAB Webots API:} Allows bidirectional communication with a simpler setup.
\end{itemize}

\section{Case Study: Inverted Pendulum}

This example is included in Webots Simulink Bridge [4]. The mathematical reference is taken from the Control Tutorials [5].

As a case study, an inverted pendulum system was modeled and simulated in Webots. The pendulum is mounted on a cart and moves along a horizontal axis. This system is a classical benchmark problem for non-linear control due to its inherent instability and nonlinear dynamics.

\begin{figure}[H]
    \centering
    \includegraphics[width=0.5\linewidth]{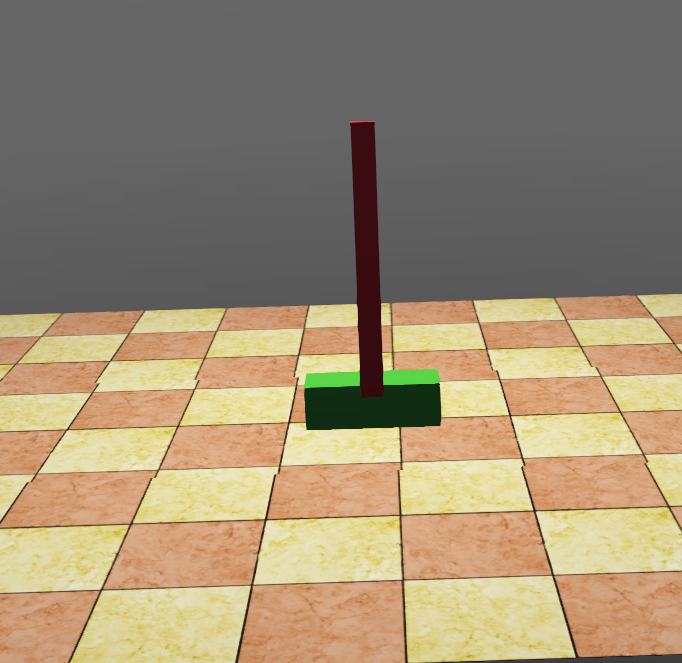}
    \caption{Webots Simulation}
    \label{fig:simulink-diagram}
\end{figure}

\subsection{Mathematical Model}

The inverted pendulum is a classical benchmark problem widely used in the fields of control systems and robotics. 
It serves as an effective testbed for evaluating system modeling, controller design, and stability analysis techniques. 
Due to its inherently unstable and nonlinear nature, the inverted pendulum requires continuous feedback control to remain balanced in the upright position.

In this study, the pendulum is mounted on a wheeled cart, and the objective is to maintain it in its vertical upright equilibrium. 
The dynamics of both the pendulum and the cart are modeled by applying Newton’s laws in the horizontal and angular directions.

According to the modeling approach presented in [5], the nonlinear equations of motion are first derived using Newton-Euler dynamics. 
These equations are then linearized under the small-angle approximation. 
Specifically, small deviations from the upright position \( \theta = \pi \) are expressed as \( \theta = \pi + \phi \), where \( \phi \approx 0 \) is assumed.

The resulting linearized model provides a foundation for the design and simulation of control strategies.

\begin{figure}[H]
    \centering
    \includegraphics[width=0.5\linewidth]{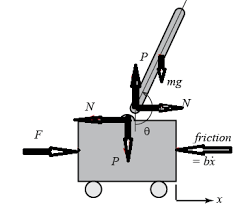}
    \caption{Inverted Pendulum }
    \label{fig:inverted-pendulum}
\end{figure}

Horizontal force balance for the cart
\begin{equation}
M\ddot{x} + b\dot{x} + N = F
\label{eq:1}
\end{equation}

Horizontal force balance for the pendulum (reaction force N)
\begin{equation}
N = m\ddot{x} - ml\ddot{\theta} \cos\theta - ml\dot{\theta}^2 \sin\theta
\label{eq:2}
\end{equation}

Equation (3) is obtained by substituting the expression for \( N \) from Equation (2) into Equation (1).
\begin{equation}
(M + m)\ddot{x} + b\dot{x} + ml\ddot{\theta} \cos\theta - ml\dot{\theta}^2 \sin\theta = F
\label{eq:3}
\end{equation}

 Force balance perpendicular to the pendulum
\begin{equation}
P\sin\theta + N\cos\theta - mg\sin\theta = ml\ddot{\theta} + m\ddot{x} \cos\theta
\label{eq:4}
\end{equation}

Moment balance about the pendulum's center of mass
\begin{equation}
-Pl\sin\theta - Nl\cos\theta = I\ddot{\theta}
\label{eq:5}
\end{equation}

By eliminating \( P \) and \( N \) from Equation (4) and Equation (5), we obtain the second nonlinear equation of motion:
\begin{equation}
(I + ml^2)\ddot{\theta} + mgl\sin\theta = -ml\ddot{x} \cos\theta
\label{eq:6}
\end{equation}

Since control and analysis techniques in this study are based on linear systems, the nonlinear equations of motion must be linearized.
Specifically, it is assumed that the pendulum oscillates around the upright equilibrium position \( \theta = \pi \). 
Small deviations from this position are represented by \( \phi \), such that \( \theta = \pi + \phi \). 
Under the small-angle assumption, the following approximations are used:

\begin{align}
\cos\theta &= \cos(\pi + \phi) \approx -1 \label{eq:7} \\
\sin\theta &= \sin(\pi + \phi) \approx -\phi \label{eq:8} \\
\dot{\theta}^2 &= \dot{\phi}^2 \approx 0 \label{eq:9}
\end{align}

Applying the approximations to the nonlinear equations yields the linearized equations of motion.
\begin{equation}
(I + ml^2)\ddot{\phi} - mgl\,\phi = ml\,\ddot{x}
\label{eq:10}
\end{equation}
\begin{equation}
(M + m)\ddot{x} + b\,\dot{x} - ml\,\ddot{\phi} = u
\label{eq:11}
\end{equation}

The nonlinear equations of motion for the inverted pendulum system are given as:

\begin{align}
(M + m)\ddot{x} + ml\ddot{\theta}\cos\theta - ml\dot{\theta}^2\sin\theta &= F \label{eq:nonlinear1} \\
l\ddot{\theta} + \ddot{x}\cos\theta - g\sin\theta &= 0 \label{eq:nonlinear2}
\end{align}

\subsubsection*{3.1.1 Transfer Function}

To obtain the transfer functions of the linearized system equations, we first apply the Laplace transform under the assumption of zero initial conditions.  
The resulting equations in the Laplace domain are:

\begin{equation}
(I + ml^2)\Phi(s)s^2 - mgl\Phi(s) = mlX(s)s^2
\tag{12}
\end{equation}

\begin{equation}
(M + m)X(s)s^2 + bX(s)s - ml\Phi(s)s^2 = U(s)
\tag{13}
\end{equation}

To find the transfer function from input \( U(s) \) to output \( \Phi(s) \), we solve Equation (12) for \( X(s) \):

\begin{equation}
X(s) = \left[ \frac{I + ml^2}{ml} - \frac{g}{s^2} \right] \Phi(s)
\tag{14}
\end{equation}

Substituting this into Equation (13) yields:

\begin{equation}
(M + m) \left[ \frac{I + ml^2}{ml} - \frac{g}{s^2} \right] s^2 \Phi(s)
+ b \left[ \frac{I + ml^2}{ml} - \frac{g}{s^2} \right] s \Phi(s)
- ml\Phi(s)s^2 = U(s)
\tag{15}
\end{equation}

Rearranging terms, the transfer function becomes:

\begin{equation}
\frac{\Phi(s)}{U(s)} =
\frac{\dfrac{ml}{q}s^2}{
s^4 + \dfrac{b(I + ml^2)}{q}s^3 - \dfrac{(M + m)mgl}{q}s^2 - \dfrac{bmgl}{q}s}
\tag{16}
\end{equation}

where

\begin{equation}
q = (M + m)(I + ml^2) - (ml)^2
\tag{17}
\end{equation}

This transfer function contains both a zero and a pole at the origin, which can be canceled, resulting in the simplified form:

\begin{equation}
P_{\text{pend}}(s) = \frac{\Phi(s)}{U(s)} = \frac{ml}{q} \cdot \frac{1}{s^2 + \dfrac{b(I + ml^2)}{mlq}s - \dfrac{(M + m)g}{q}} \quad \left[ \frac{\text{rad}}{N} \right]
\tag{18}
\end{equation}

Similarly, the transfer function from \( U(s) \) to the cart position output \( X(s) \) is given by:

\begin{equation}
P_{\text{cart}}(s) = \frac{X(s)}{U(s)} = \frac{(I + ml^2)s^2 - mgl}{s^4 + \dfrac{b(I + ml^2)}{q}s^3 - \dfrac{(M + m)mgl}{q}s^2 - \dfrac{bmgl}{q}s} \quad \left[ \frac{m}{N} \right]
\tag{19}
\end{equation}

Where:
\begin{itemize}
  \item \( x \): Horizontal position of the cart
  \item \( \theta \): Angle of the pendulum from the vertical (upward is \( \theta = 0 \))
  \item \( M \): Mass of the cart
  \item \( m \): Mass of the pendulum
  \item \( l \): Length to the pendulum center of mass
  \item \( g \): Gravitational acceleration
  \item \( F \): Horizontal force applied to the cart
\end{itemize}

Solving the equations, we can obtain the state-space representation.

\subsection{State-Space Representation}

Defining the state vector as \( \mathbf{x} = [x, \dot{x}, \theta, \dot{\theta}]^T \), the linearized system can be written in state-space form:
\[
\begin{bmatrix}
\dot{x}\\[6pt]
\ddot{x}\\[6pt]
\dot{\phi}\\[6pt]
\ddot{\phi}
\end{bmatrix}
=
\begin{bmatrix}
0 & 1 & 0 & 0 \\[6pt]
-\displaystyle \frac{\bigl(l + m'^2\bigr)\,b}{I\,(M + m) \;+\; M\,m'\,l^2} 
& 0 
& \displaystyle \frac{m'^2\,g}{I\,(M + m) + M\,m'\,l^2} 
& 0 \\[6pt]
0 & 0 & 0 & 1 \\[6pt]
-\displaystyle \frac{m\,l\,b}{I\,(M + m) + M\,m'\,l^2}
& 0 
& \displaystyle \frac{m\,g\,(M + m)}{I\,(M + m) + M\,m'\,l^2}
& 0
\end{bmatrix}
\begin{bmatrix}
x\\[4pt]
\dot{x}\\[4pt]
\phi\\[4pt]
\dot{\phi}
\end{bmatrix}
\;+\;
\begin{bmatrix}
0 \\[6pt]
\displaystyle \frac{\,l + m'^2\,}{I\,(M + m) + M\,m'\,l^2} \\[6pt]
0 \\[6pt]
\displaystyle \frac{m\,l}{I\,(M + m) + M\,m'\,l^2}
\end{bmatrix}
\,u
\]

\[
y \;=\;
\begin{bmatrix}
1 & 0 & 0 & 0\\[4pt]
0 & 1 & 0 & 0
\end{bmatrix}
\begin{bmatrix}
x\\[4pt]
\dot{x}\\[4pt]
\phi\\[4pt]
\dot{\phi}
\end{bmatrix}
\;+\;
\begin{bmatrix}
0\\
0
\end{bmatrix}
\,u
\]

\subsection{Control Implementation}

In Simulink, a PID controller was implemented to stabilize the pendulum. The angle \( \theta \) was measured via a simulated Position Sensor in Webots and sent to Simulink through a binding interface. Based on the feedback, control inputs (force \( F \)) were computed and transmitted back to Webots for real-time actuation.

Simulation results demonstrated that the pendulum remained upright under various disturbances, validating the robustness of the control strategy within the Webots–Simulink architecture.

\subsection{Simulink Control Diagram}

We create the control architecture of our robot's modeling with State Space Model without Webots.
Using PID Tuning to determine the PID coefficients, we use our coefficients for the Webots Simulink Bridge.

\begin{figure}[H]
    \centering
    \includegraphics[width=0.45\linewidth]{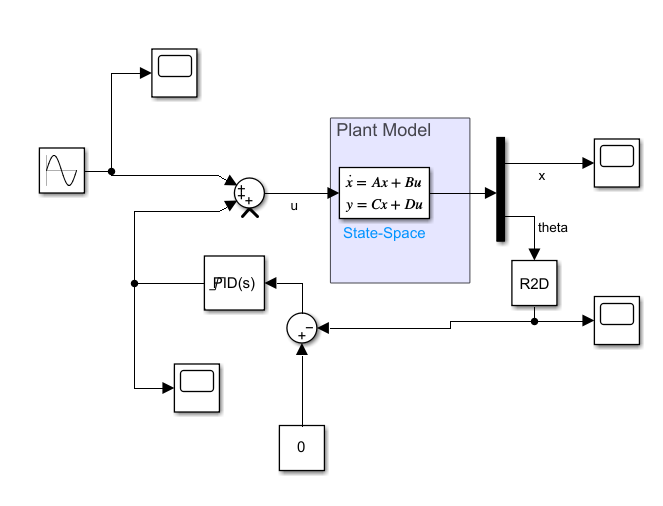}
    \caption{Simulink Control Diagram}
    \label{fig:simulink-diagram}
\end{figure}

\subsection{Simulink-Webots Control Diagram}

The blocks created provide the Webots Simulink Relationship. In the Control section of these blocks, we determine the control architecture of our robots, and in the Send Webots block, we write information to our robot in the webots. It is printed without using things like UDP/TCP. This allows real-time operations to be performed. It is shown in Figure 4.

\begin{figure}[H]
    \centering
    \includegraphics[width=1\linewidth]{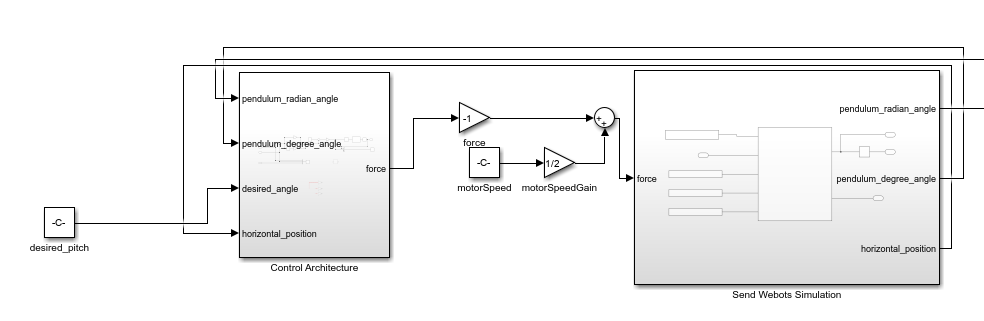}
    \caption{Webots Control Diagram}
    \label{fig:simulink-diagram}
\end{figure}

We ensure that the Pendulum remains upright by using the specified floor numbers in the Webots Simulink Control Block. It is shown in Figure 5. With the Low Pass Filter, smoother transitions are made for the printed data. 

\begin{figure}[H]
    \centering
    \includegraphics[width=1\linewidth]{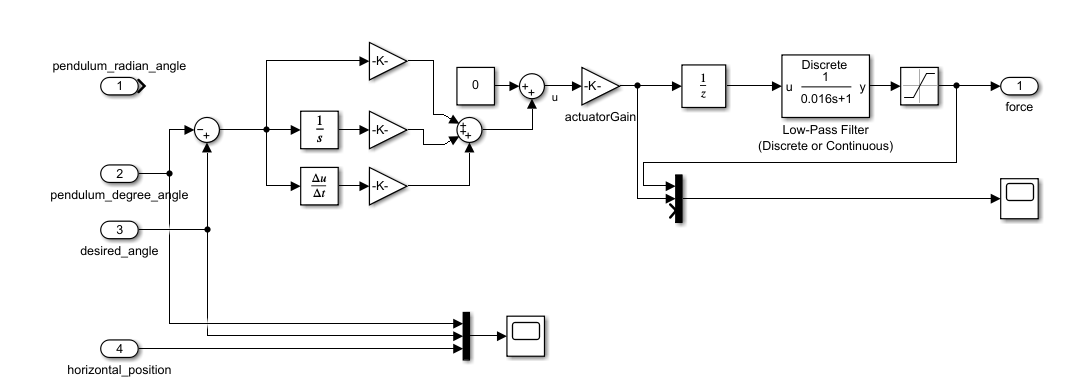}
    \caption{Control Architecture Diagram}
    \label{fig:simulink-diagram}
\end{figure}

\subsection{Outputs}

The angle \( \theta \) that we read from the Inverted Pendulum shown in Figure 6 is the degree. The image shown in Figure 7 is the data written to the actuator for the angle \( \theta \) that was read.

\begin{figure}[H]
    \centering
    \includegraphics[width=0.75\linewidth]{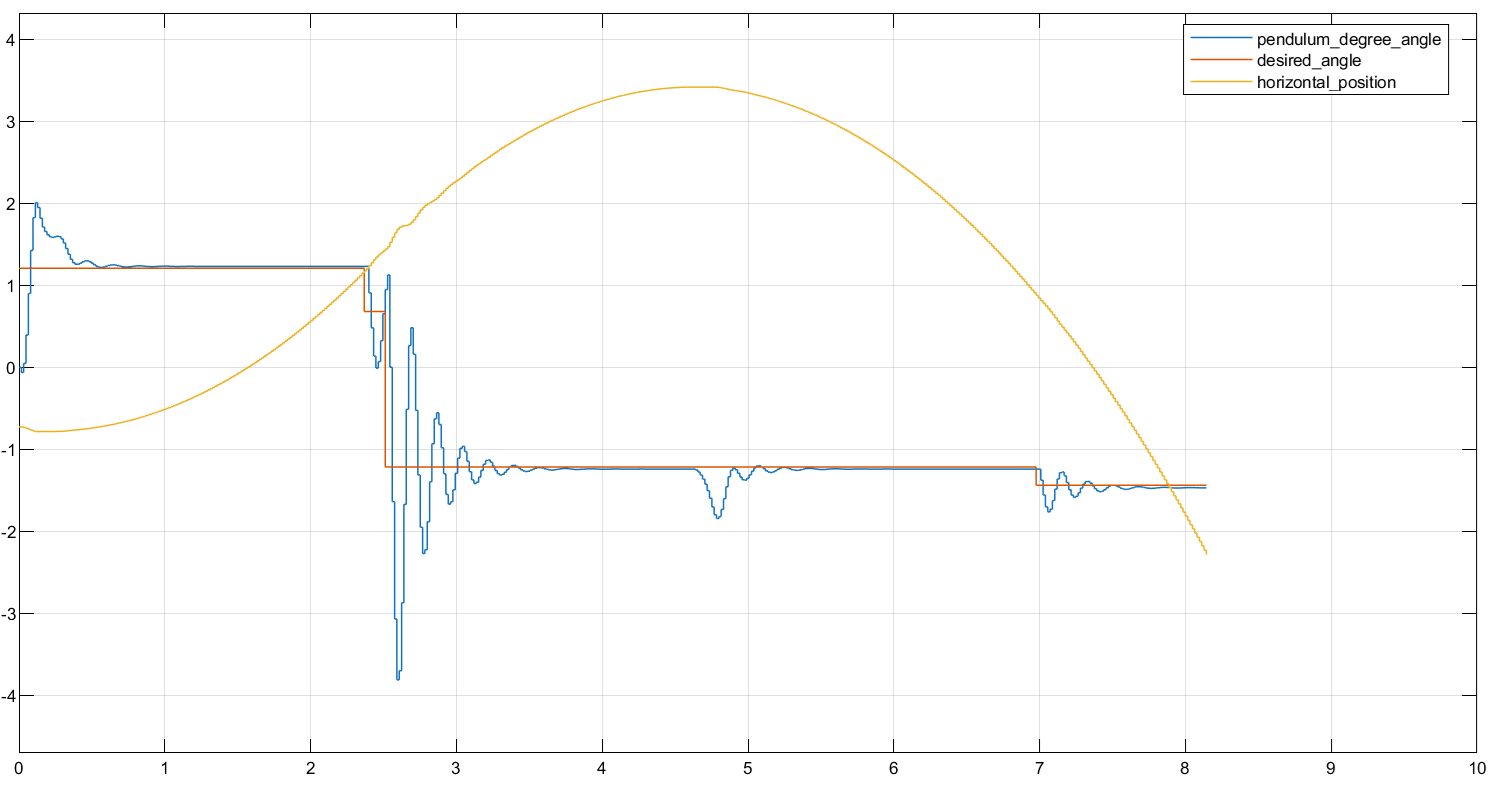}
    \caption{Pendulum (Sensor Value) and Desired (Input) Angle}
    \label{fig:simulink-diagram}
\end{figure}

\begin{figure}[H]
    \centering
    \includegraphics[width=0.75\linewidth]{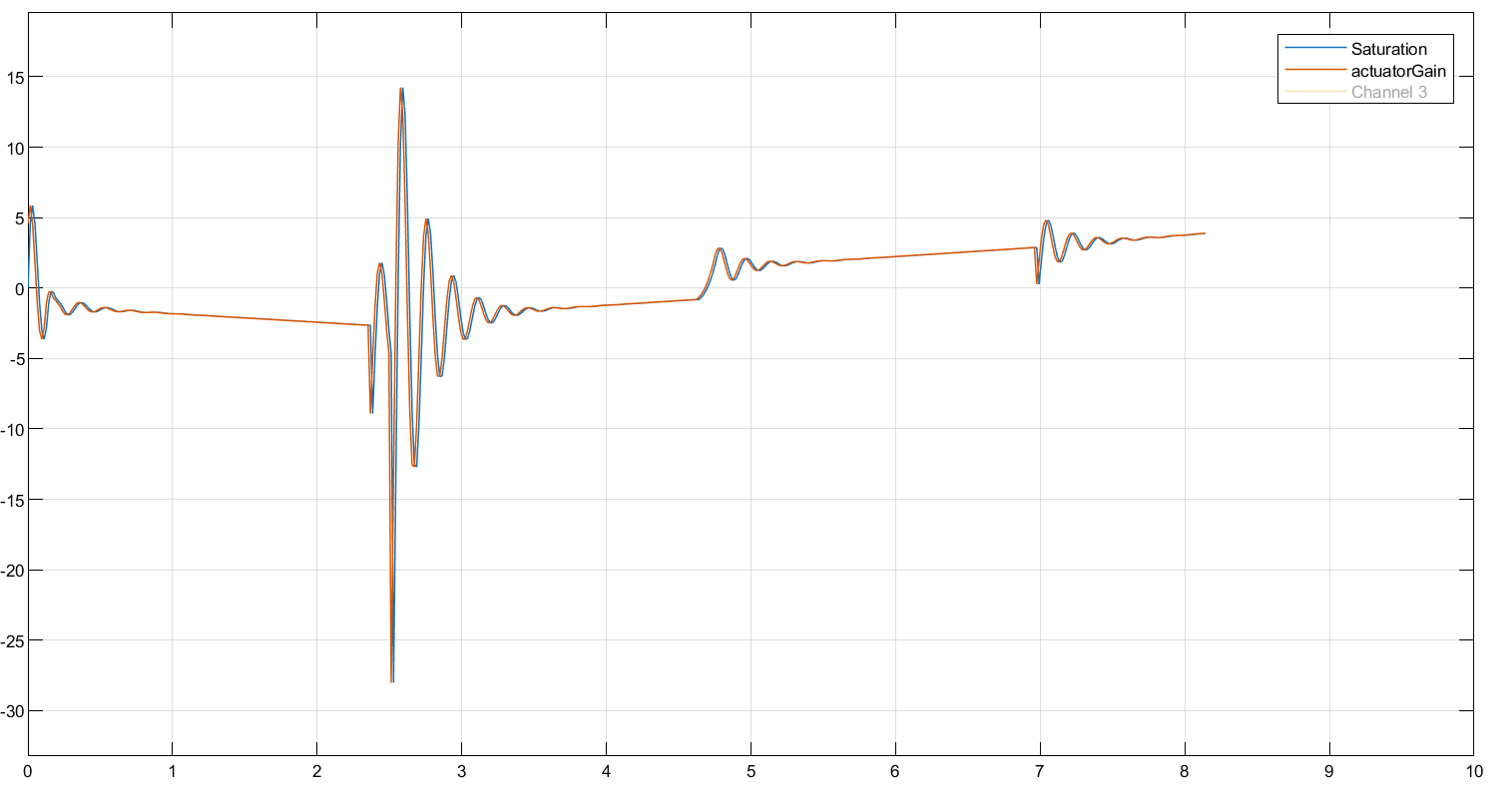}
    \caption{Actuator Outputs}
    \label{fig:simulink-diagram}
\end{figure}

\section{Contributions and Limitations}
\textbf{Contributions:}
\begin{itemize}
  \item Rapid prototyping and testing of controllers
  \item Realistic physical simulation
  \item Modular and flexible design
\end{itemize}

\textbf{Limitations:}
\begin{itemize}
  \item Communication latency in high-frequency control loops
  \item Complexity in synchronization between two platforms
\end{itemize}

\section{Conclusion}
The integration of Webots and Simulink could pave the way for the development of control algorithms by providing a multi-factor architecture for the development of robotic control systems.

\begin{figure}[H]
    \centering
    \includegraphics[width=1\linewidth]{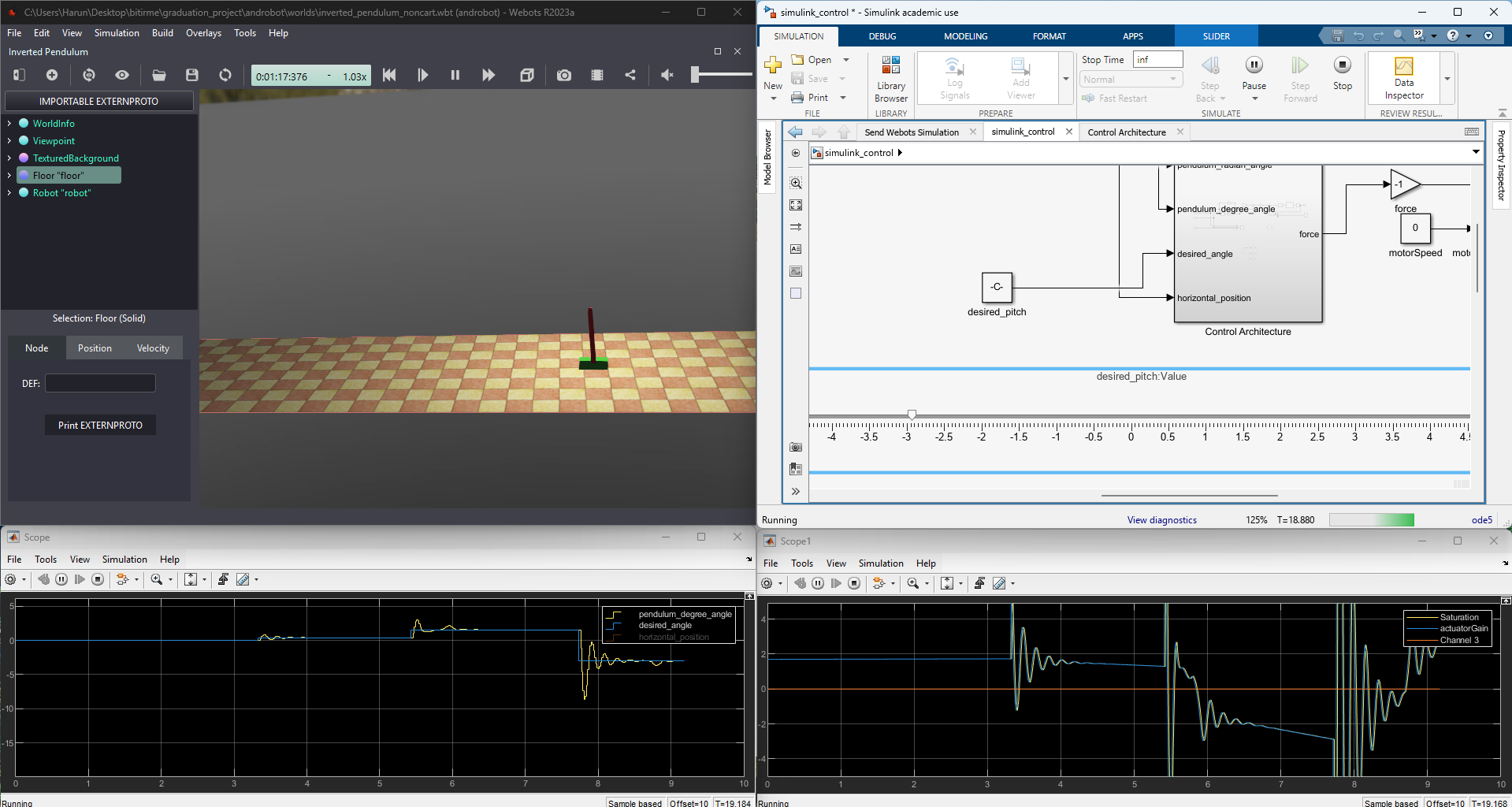}
    \caption{Webots and Simulink Couple}
    \label{fig:enter-label}
\end{figure}

\section*{Acknowledgment}
We have developed this Tool in MAGLEV Lab. and also we would like to thank Prof. Kadir ERKAN from Yıldız Technical University for his consultancy and valuable guidance throughout the study.\\

\end{document}